\documentclass{article}

\usepackage[square,numbers]{natbib}

\usepackage[preprint]{neurips_2020}

\usepackage[utf8]{inputenc} % allow utf-8 input
\usepackage[T1]{fontenc}    % use 8-bit T1 fonts
\usepackage{hyperref}       % hyperlinks
\usepackage{url}            % simple URL typesetting
\usepackage{booktabs}       % professional-quality tables
\usepackage{amsfonts}       % blackboard math symbols
\usepackage{nicefrac}       % compact symbols for 1/2, etc.
\usepackage{microtype}      % microtypography
\newcommand*\samethanks[1][\value{footnote}]{\footnotemark[#1]}
\usepackage{wrapfig}

\usepackage{caption}
\usepackage{subcaption}
\usepackage{tabularx}
\usepackage{todonotes}
\usepackage{amsmath}
\usepackage[group-separator={,}]{siunitx}

\usepackage{dutchcal}

\AtBeginDocument{%
  \providecommand\BibTeX{{%
    \normalfont B\kern-0.5em{\scshape i\kern-0.25em b}\kern-0.8em\TeX}}}

\title{SIGN: Scalable Inception Graph Neural Networks}

\author{%
  Fabrizio Frasca\thanks{Equal contribution} \\
   Twitter / Imperial College London \\
   United Kingdom\\
   \And
  Emanuele Rossi\samethanks \\
   Twitter / Imperial College London \\
   United Kingdom\\   
   \And
   Davide Eynard \\
   Twitter \\
   United Kingdom\\   
   \AND
   Ben Chamberlain \\
   Twitter \\
   United Kingdom\\   
   \And
   Michael M. Bronstein \\
   Twitter / Imperial College London\\
   United Kingdom\\   
   \And
   Federico Monti \\
   Twitter \\
   United Kingdom\\   
}

\begin{document}
\hspace{-5cm}

\maketitle

\begin{abstract}
      % maybe this first sentence can be shortened
    Graph representation learning has recently been applied to a broad spectrum of problems ranging from computer graphics and chemistry to high energy physics and social media. 
    The popularity of graph neural networks has sparked interest, both in academia and in industry, in developing methods that scale to very large graphs such as Facebook or Twitter social networks. In most of these approaches, the computational cost is alleviated by a sampling strategy retaining a subset of node neighbors or subgraphs at training time.
    In this paper we propose a new, efficient and scalable graph deep learning architecture which sidesteps the need for graph sampling by using graph convolutional filters of different size that are amenable to efficient precomputation, allowing extremely fast training and inference. 
    Our architecture allows using different local graph operators (e.g. motif-induced adjacency matrices or  Personalized Page Rank diffusion matrix) to best suit the task at hand.
    We conduct extensive experimental evaluation on various open benchmarks and show that our approach is competitive with other state-of-the-art architectures, while requiring a fraction of the training and inference time. Moreover, we obtain state-of-the-art results on \texttt{ogbn-papers100M}, the largest public graph dataset, with over 110 million nodes and 1.5 billion edges.
\end{abstract}

%%
%% The code below is generated by the tool at http://dl.acm.org/ccs.cfm.
%% Please copy and paste the code instead of the example below.
%%

%%
%% Keywords. The author(s) should pick words that accurately describe
%% the work being presented. Separate the keywords with commas.
% TODO

%% A "teaser" image appears between the author and affiliation
%% information and the body of the document, and typically spans the
%% page.

%%
%% This command processes the author and affiliation and title
%% information and builds the first part of the formatted document.
\maketitle

\section{Introduction}

Deep learning on graphs, also known as {\em geometric deep learning} (GDL) \cite{7974879} or {\em graph representation learning} (GRL) \cite{hamilton2017representation,battaglia2018relational,zhang2018deep}, has emerged in a matter of just a few years from a niche topic to one of the most prominent fields in machine learning. 
Graph deep learning models have recently scored successes in various applications relying on modeling relational data, see e.g. \cite{zhang2018link,qi2018learning,Monti2016GeometricDL,choma2018graph,NIPS2015_5954,pmlr-v70-gilmer17a,parisot2018disease,zitnik2018modeling,veselkov2019hyperfoods,gainza2019deciphering,rossi2019ncrna,monti2019fake}. % to mention a few. 
Graph Neural Networks (GNNs) seek to generalize classical convolutional architectures (CNNs) to graph-structured data, 
%
%A wide variety of convolution-like operations have been developed on graphs
with a wide variety of convolution-like operations available in the literature \cite{scarselli2008graph,defferrard2016convolutional,DBLP:conf/nips/AtwoodT16,DBLP:conf/icml/NiepertAK16,DBLP:conf/cvpr/SimonovskyK17,Monti2016GeometricDL,kipf2016semi,pmlr-v97-wu19e,DBLP:conf/iclr/VelickovicCCRLB18,GraphSAGE}. 
%including ChebNet \cite{defferrard2016convolutional}, MoNet \cite{Monti2016GeometricDL}, %GCN \cite{kipf2016semi}, S-GCN \cite{pmlr-v97-wu19e}, GAT %\cite{DBLP:conf/iclr/VelickovicCCRLB18}, and GraphSAGE \cite{GraphSAGE} 
%(we refer the reader to recent review papers \cite{7974879,hamilton2017representation,battaglia2018relational,zhang2018deep} for a comprehensive overview of deep learning on graphs and its mathematical underpinnings). 
%
%
%
%
%In particular, graph convolutional networks (GCN) \cite{kipf2016semi} and their more recent variant S-GCN \cite{pmlr-v97-wu19e} apply a shared node-wise linear transformation of the node features, followed by one or more iterations of diffusion on the graph. 
%

Until recently, most of the research in the field has focused on small-scale datasets,  %(CORA \cite{sen2008collective} with only $\sim 5$K nodes still being among the most widely used), 
and relatively little effort has  been devoted to scaling these methods to web-scale graphs such as the Facebook or Twitter social networks.
Scaling is a major challenge precluding the wide application of graph deep learning methods in industrial settings. Compared to Euclidean neural networks where the training loss can be decomposed into individual samples and computed independently, graph convolutional networks diffuse information between nodes along the edges of the graph, making the loss computation interdependent for different nodes. 
Furthermore, in typical graphs the number of nodes grows exponentially with the increase of the filter receptive field, incurring significant computational and memory complexity. 
So far, various 
{\em graph sampling} approaches  \cite{GraphSAGE, pinsage, fastgcn, adaptive-sampling, stochastic-training,Chiang:2019:CEA:3292500.3330925,DBLP:journals/corr/abs-1907-04931,zou2019layer} have been proposed as a way to alleviate the cost of training graph neural networks by selecting a small number of neighbors that reduce the computational and memory complexity. %
%Such methods can potentially scale to web-size graphs \cite{pinsage}. 
%GraphSAGE \cite{GraphSAGE} uniformly randomly samples the neighborhood of a given node. PinSAGE \cite{pinsage} uses random walks to improve the quality of such approximation. ClusterGCN \cite{Chiang:2019:CEA:3292500.3330925} clusters the graph and enforces diffusion of information only within the computed clusters. GraphSAINT \cite{DBLP:journals/corr/abs-1907-04931} uses  unbiased estimators of neighborhood graphs. %They propose multiple methods to sample minibatch subgraphs during training while using normalization technique to eliminate bias.

In this paper, we take a different approach for scalable deep learning on graphs. We propose \textit{SIGN}, a simple scalable Graph Neural Network architecture inspired by the inception module  \cite{szegedy2015going, anees2019inception}. \textit{SIGN} combines graph convolutional filters of different types and sizes that are amenable to efficient precomputation, allowing extremely fast training and inference with complexity independent of the graph structure. Our architecture is able to scale to web-scale graphs without resorting to any sample technique and retaining sufficient expressiveness for effective learning: while being faster in training and, especially, inference (even one order of magnitude speedup), by employing \textit{SIGN} with only one graph convolutional layer we are able to achieve results on par with state-of-the-art on several large-scale graph learning datasets. In particular, \textit{SIGN} obtains state-of-the-art results on \texttt{ogbn-papers100M}, the largest public graph learning benchmark, with over 110 million nodes and 1.5 billion edges.
%
% Our architecture generalizes several previous methods such as GCN \cite{kipf2016semi}, S-GCN \cite{pmlr-v97-wu19e}, ChebNet \cite{defferrard2016convolutional}, and MotifNet \cite{monti2018motifnet}, and is compatible with various graph sampling approaches.
%
% While being faster in training and, especially, inference (even one order of magnitude speedup), by employing SIGN with only one graph convolutional layer we are able to achieve results on par with state-of-the-art on large-scale graph learning datasets
% An important observation of our paper is that employing SIGN with only one graph convolutional layer, we are able to achieve results on par with or superior to the state-of-the-art, while being faster in training and, especially, inference (even one order of magnitude speedup). We provide extensive experimental validation of this claim on large-scale graph learning datasets. SIGN obtains state-of-the-art results on OGBN-papers100M, the largest public graph learning benchmark, with over 110 million nodes and 1.5 billion edges.

These results raise the important question on when deep graph neural network architectures are useful, especially when scalability is an important requirement, as in large-scale industrial systems. 
Significant effort has recently been devoted to methods allowing to design deep Graph Neural Networks with many graph convolutional layers \cite{jk,gong2020geometrically,li2019deepgcns,zhao2019pairnorm,rong2019dropedge}, which otherwise appear difficult to train \cite{li2018adaptive,klicpera2018predict,wu2020comprehensive}. 
While there is strong evidence in favor of depth on geometric graphs \cite{he2016deep,li2019deepgcns,gong2020geometrically}, there has been almost no gain from depth on general irregular graphs like `small-world' networks \cite{shchur2018pitfalls,rong2019dropedge}. 
Given the abundance of such graphs e.g. in social network applications, it is important to take a step back and deliberate if deep architectures are the right approach.
We conjecture that deep graph learning architectures are not useful for general irregular graphs and argue that future research in the field should focus on designing local more expressive operators \cite{barbarossa2019topological,monti2018motifnet,flam2020neural} rather than going deeper.

%We provide extensive experimental validation showing that, despite its simplicity, our approach produces comparable results to state-of-the-art architectures on a variety of large-scale graph datasets while being significantly faster (orders of magnitude) in training and inference.  

\section{Background and Related Work} \label{sec:background}

\subsection{Deep learning on graphs}

The goal of graph representation learning is to construct an embedding representing the structure of the graph and the data thereon.
In node-wise prediction problems, we distinguish between {\bf Transductive} setting, which assumes that the entire graph is known, and thus the same graph is used during training and testing (albeit different nodes are used for training and testing), and {\bf Inductive} setting, in which training and testing are performed on different graphs.
A typical graph neural network architecture consists of graph {\bf Convolution-like operators} (discussed in Section~\ref{sec:convo}) performing local aggregation of features by means of message passing with the neighbor nodes, and possibly {\bf Pooling} amounting to fixed \cite{dhillon2007weighted} or learnable \cite{ying2018hierarchical,bianchi2019mincut} graph coarsening. 
Additionally, graph {\bf Sampling} schemes (detailed in Section~\ref{sec:sampling}) can be employed on large-scale graphs to reduce the computational complexity. 

\subsection{Basic notions}

% Graph and notation, GCNs and so on
Let $\mathcal{G} = (\mathcal{V} = \{1,\hdots, n\}, \mathcal{E}, \mathbf{W} )$ be an undirected weighted graph, represented by the symmetric $n\times n$ {\em adjacency matrix} $\mathbf{W}$, where $w_{ij} > 0$ if $(i,j) \in \mathcal{E}$ and zero otherwise. 
The diagonal {\em degree matrix} $\mathbf{D} =  \mathrm{diag}(\sum_{j=1}^n w_{1j}, \hdots, \sum_{j=1}^n w_{nj})$ represents the number of neighbors of each node. 
We further assume that each node is endowed with a $d$-dimensional feature vector and arrange all the node features as rows of the $n\times d$-dimensional matrix $\mathbf{X}$. 
We denote by $\mathbf{A} = \mathbf{D}^{-1/2}\mathbf{W} \mathbf{D}^{-1/2}$ the normalized adjacency matrix. The normalized {\em graph Laplacian} is an $n\times n$ positive semi-definite matrix $\boldsymbol{\Delta} = \mathbf{I} - \mathbf{D}^{-1/2}\mathbf{W} \mathbf{D}^{-1/2}$. %, where 

\subsection{Convolution-like operators on graphs} \label{sec:convo}

%{\bf Spectral methods. \hspace{3mm}}
\paragraph{Spectral methods}
Bruna et al. \cite{bruna2013spectral} used the analogy between the eigenvectors of the graph Laplacian and the Fourier transform to generalize convolutional neural networks (CNN) \citep{lecun1989backpropagation} to graphs. 
Among the key drawbacks of this approach is a high (at least $\mathcal{O}(n^2)$ complexity), 
large ($\mathcal{O}(n)$) number of filter parameters,  
no spatial localization, and no generalization of filters across graphs. Furthermore, the method explicitly assumes the underlying graph to be undirected, in order for the Laplacian to be a symmetric matrix with orthogonal eigenvectors. 

% {\bf ChebNet. \hspace{3mm}}
\paragraph{ChebNet}
A way to address the shortcomings of \cite{bruna2013spectral} is to model the filter as a transfer function $\hat{g}(\lambda)$, applied to the Laplacian as $\hat{g}(\boldsymbol{\Delta}) = \boldsymbol{\Phi}\hat{g}(\boldsymbol{\Lambda})\boldsymbol{\Phi}^\top$. Filters computed in this manner are stable under graph perturbations \cite{levie2019transferability}. 
In the case when $\hat{g}$ is expressed as simple matrix-vector operations (e.g. a polynomial \cite{defferrard2016convolutional} or rational function \cite{levie2018cayleynets}), the eigendecomposition of the Laplacian  can be avoided altogether. 
A particularly simple choice is a polynomial spectral filter $\hat{g}(\lambda) = \sum_{k=0}^r \theta_k \lambda^k$ of degree $r$, allowing the convolution to be computed entirely in the spatial domain as %\vspace{-1mm}
\begin{equation}
    \mathbf{Y} = \hat{g}(\boldsymbol{\Delta}) \mathbf{X} = \sum_{k=0}^r \theta_k \boldsymbol{\Delta}^k \mathbf{X}.
    \label{eqn:cheby}
\end{equation}
\noindent with  $\mathcal{O}(r)$ parameters $\theta_0, \hdots, \theta_r$, does not require explicit multiplication by $\boldsymbol{\Phi}$, and has a compact support of $r$ hops in the node domain. %  (due to the fact that $\boldsymbol{\Delta}^k$ affects only neighbors within $k$-hops). 
Though originating from a spectral construction, the resulting filter is an operation in the node domain amounting to a successive aggregation of features in the neighbor nodes,  
%
%Second, using recursively-defined Chebyshev polynomials,  
%$T_{j+1}(\lambda) = 2\lambda T_{j}(\lambda) - T_{j-1}(\lambda)$ with $T_1(\lambda) = \lambda$ and $T_0(\lambda)=1$, 
%the computation 
which can be performed with complexity  $\mathcal{O}(|\mathcal{E}|r) \approx \mathcal{O}(nr)$. % for sparsely-connected graphs. 
The polynomial filters can be combined with non-linearities, concatenated in multiple layers, and interleaved with pooling layers based on graph coarsening \cite{defferrard2016convolutional}. 
The Laplacian in~(\ref{eqn:cheby}) can be replaced with other operators that diffuse information across neighbor nodes, e.g. the simple or normalized adjacency matrix, without affecting performance. 

% {\bf GCN. \hspace{3mm}}
\paragraph{GCN}
In the case $r=1$, equation~(\ref{eqn:cheby}) reduces to computing $(\mathbf{I} + \mathbf{D}^{-1/2}\mathbf{W} \mathbf{D}^{-1/2}) \mathbf{X}$, which can be interpreted as a combination of the node features and the neighbors filtered features. 
Kipf and Welling \cite{kipf2016semi} proposed a model of graph convolutional networks (\textit{GCN}) combining node-wise and graph diffusion operations: 
\begin{equation} \label{eq:gcn}
    \mathbf{Y} = \tilde{\mathbf{D}}^{-1/2} \tilde{\mathbf{W}}\tilde{\mathbf{D}}^{-1/2} \mathbf{X}\boldsymbol{\Theta} = \tilde{\mathbf{A}}\mathbf{X}\boldsymbol{\Theta}. 
\end{equation}
Here $\tilde{\mathbf{W}} = \mathbf{I} + \mathbf{W}$ is the adjacency matrix with self-loops, $\tilde{\mathbf{D}} = \mathrm{diag}(\sum_{j=1}^n \tilde{w}_{1j},\hdots, \sum_{j=1}^n \tilde{w}_{nj})$ is the respective degree matrix, and $\boldsymbol{\Theta}$ is a matrix of learnable parameters. 

% {\bf S-GCN. \hspace{3mm}}
\paragraph{S-GCN}
Stacking $L$ \textit{GCN} layers with element-wise non-linearity $\sigma$ and a final layer for node classification with activation $\xi$ (e.g. softmax or sigmoid), it is possible to obtain filters with larger receptive fields on the graph nodes, 
\begin{equation} \nonumber 
   \mathbf{Y} = \xi( \tilde{\mathbf{A}} \cdots \sigma(\tilde{\mathbf{A}}\mathbf{X}\boldsymbol{\Theta}^{(1)}) \cdots \boldsymbol{\Theta}^{(L)}).
\end{equation}
\noindent Wu et al. \cite{pmlr-v97-wu19e} argued that graph convolutions with large filters is practically equivalent to multiple convolutional layers with small filters and showed that all but the last non-linearities can be removed without practically harming the performance, resulting in the \emph{simplified GCN} (\textit{S-GCN}) model, 
\begin{equation} \label{eq:sgcn}
    \mathbf{Y} = \xi( \tilde{\mathbf{A}}^L \mathbf{X}\boldsymbol{\Theta}^{(1)} \cdots \boldsymbol{\Theta}^{(L)}) = \xi( \tilde{\mathbf{A}}^L \mathbf{X}\boldsymbol{\Theta}). 
\end{equation}

% {\bf MotifNet. \hspace{3mm}}
\paragraph{MotifNet}
Monti et al. \cite{monti2018motifnet} used adjacency matrices with weights proportional to the count of simple subgraphs (motifs) on edges in order to account for higher order structures. Related ideas have been explored using higher-order Laplacians on simplicial complexes \cite{barbarossa2019topological}.

\subsection{Graph sampling} \label{sec:sampling}

For Web-scale graphs such as Facebook or Twitter that typically have $n = 10^8 \sim 10^9$ nodes and $|\mathcal{E}| = 10^{10} \sim 10^{11}$ edges, the diffusion matrix cannot be stored in memory for training, making the straightforward application of graph neural networks impossible. 
Graph sampling has been shown to be a successful technique to scale GNNs to large graphs by approximating local connectivity with subsampled versions which are amenable for computation. 

\paragraph{Node-wise sampling} These strategies perform graph convolutions on \emph{partial} node neighborhoods to reduce computational and memory complexity, and are coupled with minibatch training, where each training step is performed only on a batch of nodes rather than on  the whole graph. A training batch is assembled by first choosing $b$ `optimization' nodes and partially expanding their corresponding neighborhoods. In a single training step, the loss is computed and optimized only for optimization nodes. 
Node-wise sampling coupled with minibatch training was first introduced in \textit{GraphSAGE} \cite{GraphSAGE} to address the challenges of scaling GNNs. PinSAGE \cite{pinsage} extended \textit{GraphSAGE} by exploiting a neighbor selection method using scores from approximations of Personalized PageRank \cite{haveliwala2003topic} via random walks. \textit{VR-GCN} \cite{stochastic-training} uses control variates to reduce the variance of  stochastic training and increase the speed of convergence with a small number of neighbors.

\paragraph{Layer-wise sampling} A characteristic of many graphs, in particular `small-world'  social networks, is the exponential growth of the neighborhood size with number of hops $L$. \cite{fastgcn, adaptive-sampling} avoid over-expansion of neighborhoods to overcome the redundancy of node-wise sampling. Nodes in each layer only have directed edges towards nodes of the next layer, thus bounding the maximum amount of computation to $\mathcal{O}(b^2)$ per layer. Moreover, sharing common neighbors prevents feature replication across the batch, drastically reducing the memory complexity during training.

\paragraph{Graph-wise sampling} In \cite{Chiang:2019:CEA:3292500.3330925, DBLP:journals/corr/abs-1907-04931}, feature sharing is further advanced: each batch consists of a connected subgraph and at each training iteration the GNN model is optimized over all nodes in the subgraph.
In \textit{ClusterGCN} \cite{Chiang:2019:CEA:3292500.3330925}, non-overlapping clusters are computed as a pre-processing step and then sampled during training as input minibatches. \textit{GraphSAINT} \cite{DBLP:journals/corr/abs-1907-04931} adopts a similar approach, while also correcting for the bias and variance of the minibatch estimators when sampling subgraphs for training. It also explores different schemes to sample the subgraphs such as a random walk-based sampler, which is able to co-sample nodes having high influence on each other and guarantees each edge has a non-negligible probability of being sampled.

\begin{table}
%\vspace{-17mm}
% \begin{table}[t!]
    \caption{Theoretical time complexity where $L_\mathrm{c}, L_\mathrm{ff}$ is the number of graph convolition and MLP layers, $r$ is the filter size, $N$ the number of nodes (in training or inference,  respectively), $|\mathcal{E}|$ the number of edges, and $d$ the feature dimensionality (assumed fixed for all layers). For \textit{GraphSAGE}, $k$ is the number of sampled neighbors per node. Forward pass complexity corresponds to an entire epoch where all nodes are seen.}
    \label{tab:complexity}
    \centering
    \begin{tabular}{| l | cc |}
    \hline
     & \textit{Preproc.} & \textit{Forward Pass} \\ \hline
    \textit{GraphSAGE}  & $\mathcal{O}(k^{L_\mathrm{c}} N)$ &  $\mathcal{O}(k^{L_\mathrm{c}}N d^2)$ \\ 
    \textit{ClusterGCN}  & $\mathcal{O}(|\mathcal{E}|)$ & $\mathcal{O}(L_\mathrm{c}|\mathcal{E}|d + L_\mathrm{ff}Nd^2)$ \\ 
    \textit{GraphSAINT} & $\mathcal{O}(kN)$ & $\mathcal{O}(L_\mathrm{c}|\mathcal{E}|d + L_\mathrm{ff} Nd^2)$ \\ 
    \hline
    \textit{SIGN-$r$} & $\mathcal{O}(r|\mathcal{E}|d)$ & $\mathcal{O}(r L_\mathrm{ff} N d^2)$ \\  
    \hline
    \end{tabular} 
\end{table}

\section{Scalable Inception Graph Neural Networks}\label{s:method}

\begin{figure}%0.5\textwidth}
    \centering
    \includegraphics[width=0.5\linewidth]{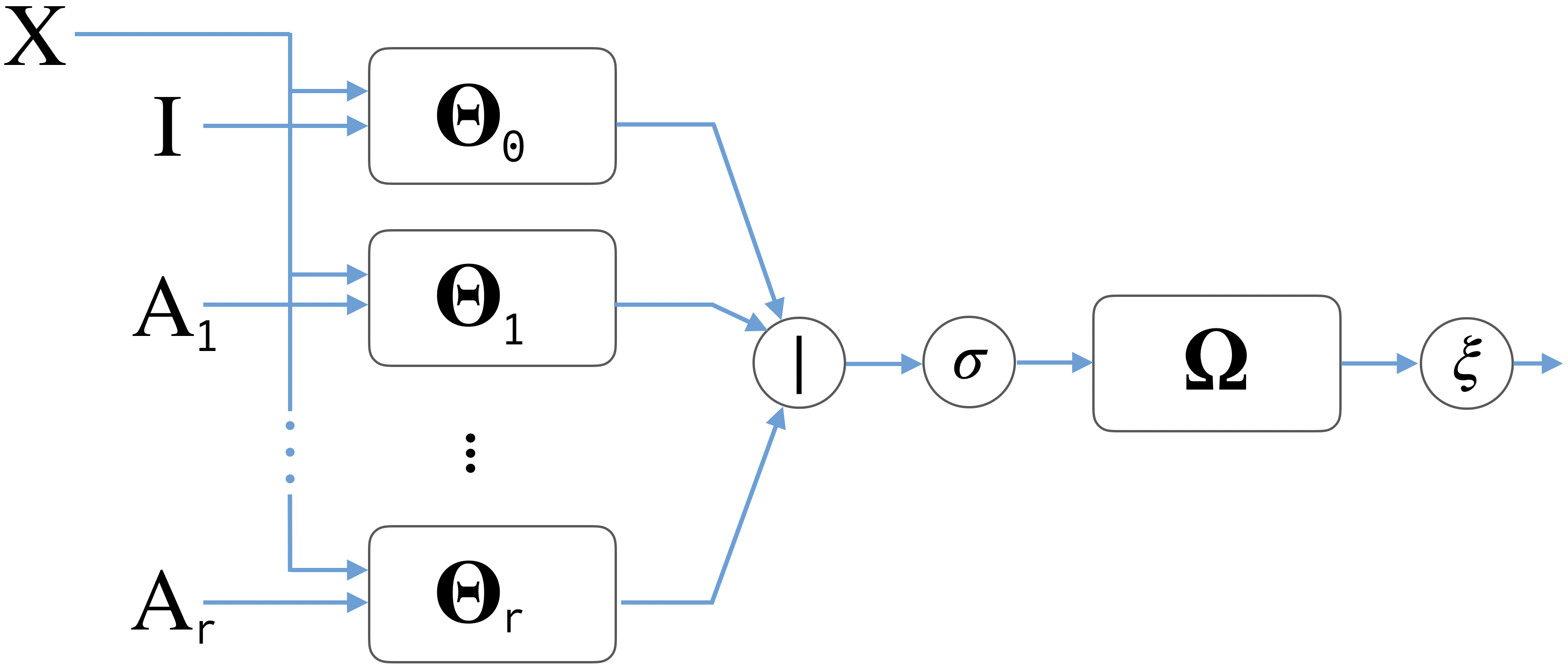}
    \caption{The \textit{SIGN} architecture for $r$ generic graph filtering operators. $\boldsymbol{\Theta}_{k}$ represents the $k$-th dense layer transforming node-wise features downstream the application of operator $k$, $\mid$ is the concatenation operation and $\boldsymbol{\Omega}$ refers to the dense layer used to compute final predictions.}
    \label{fig:sign-architecture}
\end{figure}

In this work we propose \textit{SIGN}, an alternative method to scale Graph Neural Networks to very large graphs.
The key building block of our architecture is a set of linear diffusion operators represented as $n\times n$ matrices $\mathbf{A}_1, \hdots, \mathbf{A}_r$, whose application to the node-wise features can be pre-computed.  
For node-wise classification tasks, our architecture has the form (Figure~\ref{fig:sign-architecture}):
\begin{eqnarray}
    \mathbf{Z} \hspace{-1mm} &=& \hspace{-1mm} \sigma\left([ 
    \mathbf{X} \boldsymbol{\Theta}_0, 
    \mathbf{A}_1 \mathbf{X} \boldsymbol{\Theta}_1,  \hdots , 
    \mathbf{A}_r \mathbf{X} \boldsymbol{\Theta}_r
    ]
    \right)  
    \label{eqn:sign_layer} \nonumber\\
    \mathbf{Y} \hspace{-1mm} &=& \hspace{-1mm} 
    \xi \left( \mathbf{Z} \boldsymbol{\Omega}
    \right),
    \label{eqn:sign}
\end{eqnarray}
\noindent where $\boldsymbol{\Theta}_0, \hdots, \boldsymbol{\Theta}_r$ and  $\boldsymbol{\Omega}$ are learnable matrices respectively of dimensions $d\times d'$ and $d'(r+1)\times c$ for $c$ classes, and $\sigma$, $\xi$ are non-linearities, the second one computing class probabilities, e.g. via softmax or sigmoid function, depending on the task at hand. We denote a model with $r$ operators by {\em SIGN-$r$}. 

A key observation is that matrix products $\mathbf{A}_1\mathbf{X}, \hdots, \mathbf{A}_r\mathbf{X}, $ in equation~(\ref{eqn:sign}) {\em do not depend} on the learnable model parameters and can be easily precomputed. For large graphs, distributed computing infrastructures such as Apache Spark can speed up computation. This effectively reduces the computational complexity of the overall model to that of a Multi-Layer Perceptron (MLP), i.e. $\mathcal{O}(rL_\mathrm{ff}Nd^2)$, where $d$ is the number of features, $N$ the number of nodes in the training/testing graph and $L_\mathrm{ff}$ is the overall number of feed-forward layers in the model. 

Table~\ref{tab:complexity} compares the complexity of our \textit{SIGN} model to the other scalable architectures \textit{GraphSAGE}, \textit{ClusterGCN}, and \textit{GraphSAINT}. While all models scale linearly w.r.t. the number of nodes $N$, we show experimentally that our model is significantly faster than all others due to the fact the forward and backward pass complexity of our model does not depend on the graph structure.
Unlike the aforementioned scalable methods, \textit{SIGN} is not based on sampling nodes or subgraphs, operations potentially introducing bias into the optimization procedure. 
%However, SIGN is fully compatible with graph sampling approaches that amount to replacing the operators $\mathbf{A}_k$ with their sampled versions.  

% {\bf Inception-like module. \hspace{3mm} }
\paragraph{Inception-like module}
Within the \textit{SIGN} framework, it is possible to choose one specific operator $\mathbf{B}$ and to define $\mathbf{A}_k = \mathbf{B}^k$ for $k=1, \hdots, r$. In this setting the proposed model is analogous to the popular {\em Inception module} \cite{szegedy2015going} for classic CNN architectures: it consists of convolutional filters of different sizes determined by the parameter $r$, where $r=0$ corresponds to $1\times1$ convolutions in the inception module (amounting to linear transformations of the features in each node without diffusion across nodes).
Owing to this analogy, we refer to our model as the \emph{Scalable Inception Graph Network} (\textit{SIGN}).
%
%We notice that one work extending the idea of an Inception module to GNNs is the one in~\cite{anees2019inception}; in this work, however, authors do not discuss the inclusion of a linear, non-diffusive term ($r=0$) which effectively accounts for a skip connection, and consider only Laplacian operators. Additionally, the focus is not on scaling the model to large graphs, but rather on capturing intra- and inter-graph structural heterogeneity.
% 
It is also easy to observe  that various graph convolutional layers can be obtained as particular settings of~(\ref{eqn:sign}). In particular, by setting the $\sigma$ non-linearity to PReLU \cite{prelu}, ChebNet, GCN, and \textit{S-GCN} can be automatically learnt if suitable diffusion operator $\mathbf{B}$ and activation $\xi$ are used (see Table \ref{tab:reduction}).
 
% {\bf Choice of the operators.\hspace{3mm} }
\paragraph{Choice of the operators}
Generally speaking, the choice of the diffusion operators jointly depends on the task, graph structure, and the features. 
In complex networks such as social graphs, operators induced by triangles or cliques might help distinguishing edges representing weak or strong ties \cite{Granovetter82thestrength}. In graphs with noisy connectivity, it was shown that diffusion operators based on Personalized PageRank (PPR) or Heat Kernel can boost performance \citep{klicpera_diffusion_2019}. In our experiments, we choose three specific types of operators: simple (normalized) adjacency, Personalized PageRank-based adjacency, and triangle-based adjacency matrices, as well as their powers. 
We denote by {\em SIGN($p$,$s$,$t$)} with $r=p+s+t$ the configuration using up to the $p$-th, $s$-th, and $t$-th power of simple GCN-normalized, PPR-based, and triangle-based adjacency matrices, respectively.
Lastly, when working on directed graphs, \textit{SIGN} can be equipped with powers of the (properly normalized) directed adjacency matrix $\mathbf{W}_d$ and its transpose $\mathbf{W}_d^\top$, in addition to the standard operators built on top of its undirected counterpart $\mathbf{W} = \frac{1}{2}(\mathbf{W}_d + \mathbf{W}_d^\top)$.

\begin{table}[t!]
    \caption{By appropriate configuration, \textit{SIGN} inception layer is able to replicate some popular graph convolutional layers. $\alpha$ represents the learnable parameter of a PReLU activation.}
    \label{tab:reduction}
    \centering
    \begin{tabular}{| l | c c c c |}
        \hline
            & $\mathbf{B}_1, \hdots, \mathbf{B}_r$
            & $\alpha$
            & $\mathbf{\Theta}_0, \hdots, \mathbf{\Theta}_r$
            & $\mathbf{\Omega}$ \\
        \hline
        \textit{ChebNet} \cite{defferrard2016convolutional}
            & $\mathbf{\Delta}, \hdots, \mathbf{\Delta}^r$
            & $1$
            & $\mathbf{\Theta}_{0}, \hdots, \mathbf{\Theta}_r $
            & $[\mathbf{I} , \hdots , \mathbf{I}]^\top$ \\
        
        \textit{GCN} \cite{kipf2016semi}
            & $r=1, \,\, \tilde{\mathbf{A}}$
            & $1$
            & $\mathbf{0}, \mathbf{\Theta}$
            & $[\mathbf{0}, \mathbf{I}]^\top$\\
        
        \textit{S-GCN} \cite{pmlr-v97-wu19e}
            & $r=1, \,\, \tilde{\mathbf{A}^L}$
            & $1$
            & $\mathbf{0}, \mathbf{\Theta}$
            & $[\mathbf{0}, \mathbf{I}]^\top$\\
        
        \hline
    \end{tabular}%\vspace{-3mm}
\end{table}

% {\bf \textit{SIGN} and \textit{S-GCN}.\hspace{3mm} }
\paragraph{SIGN and S-GCN}
\textit{SIGN} and \textit{S-GCN} are the only graph neural models which are inherently `shallow': contrary to standard GNN architectures, graph convolutional layers are not sequentially stacked, but either collapsed into a single linear filtering operation (S-GCN) or applied in parallel to obtain multi-scale node representations capturing diverse connectivity patterns depending on the chosen operators (\textit{SIGN}). This is the crucial feature that allows these models to naturally scale their training and inference to graphs of any size and family given that all graph operations can be conveniently pre-computed. While we notice that \textit{S-GCN} can be considered as a specific configuration of \textit{SIGN}-$1$ (it is sufficient to choose $\mathbf{A}_1 = \mathbf{\tilde{A}}^L$ and to constrain $\mathbf{\Theta}_0$ to $\mathbf{0}$, see Table~\ref{tab:reduction}), we remark the fact that the more general \textit{SIGN} architecture easily allows to incorporate more expressivity via parallel application of several, possibly, domain-specific, operators. We experimentally show this in the following section where we demonstrate that the \textit{S-GCN} paradigm is indeed too limiting and that a more expressive \textit{SIGN} model is not only able to perform on par with `deeper' sampling-based models, but also to achieve state-of-the-art results on the largest publicly available graph learning benchmark.

\section{Experiments} \label{sec:experiments}

% {\bf Datasets \hspace{3mm}}
\paragraph{Datasets}
We evaluated the proposed method on node-wise classification tasks, both in transductive and inductive settings. 
Inductive experiments are performed using four datasets: \texttt{Reddit} \cite{GraphSAGE}, \texttt{Flickr}, \texttt{Yelp} \cite{DBLP:journals/corr/abs-1907-04931}, and \texttt{PPI} \cite{Zitnik_2017}. To date, these are the largest graph learning inductive node classification benchmarks available in the public domain. Related tasks are multiclass node-wise classification for \texttt{Reddit} and \texttt{Flickr} and multilabel classification for \texttt{Yelp} and \texttt{PPI}. %: in the former, the task is to predict communities of online posts based on user comments; in the latter, the task is image categorization based on the description and common properties of online images. Yelp and PPI are multilabel classification problems: the objective of the former is to predict business attributes based on customer reviews, while the latter task consists of predicting protein functions from the interactions of human tissue proteins. 
Transductive experiments were performed on the new \texttt{ogbn-products} and \texttt{ogbn-papers100M} datasets \cite{ogb2020}. The former represents an Amazon product co-purchasing network \cite{Bhatia16} where the task is to predict the category of a product in a multi-class classification setup. The latter represents a directed citation network of $ \sim 111$ million academic papers, where the task is to leverage information from the entire citation network to infer the labels (subject areas) of a smaller subset of ArXiv papers. Overall, this dataset is orders-of-magnitude larger than any existing node classification dataset and is therefore the most important testbed for the scalability of \textit{SIGN} and related methods.

Furthermore, we also test the scalability of our method on \texttt{Wikipedia} links \cite{Konect2017wikipedia}, a large-scale network of links between articles in the English version of Wikipedia. 

Statistics for all the datasets are reported in Table~\ref{tab:dataset-statistics-ind}.

% {\bf Setup. \hspace{3mm}} 
\paragraph{Setup}
We tested several {\em SIGN}$(p,s,t)$ configurations, with $p$ the maximum power of the GCN-normalized adjacency matrix, $s$ that of a random-walk normalized PPR diffusion operator \cite{klicpera_diffusion_2019}, and $t$ that of a row-normalized triangle-induced adjacency matrix \cite{monti2018motifnet}, with weights proportional to edge occurrences in closed triads. PPR-based operators are computed from a symmetrically normalized adjacency transition matrix in an approximated form, with a restart probability of $\alpha=0.01$ for inductive datasets and $\alpha=0.05$ in the transductive case.
To allow for larger model capacity in the inception modules and in computing final model predictions, we replace the single-layer projections performed by $\mathbf{\Theta}_{i}$ and $\mathbf{\Omega}$ modules with multiple feedforward layers. Model parameters are found by minimizing the cross-entropy loss via minibatch gradient descent with the Adam optimizer \cite{adam}. Early stopping is applied with a patience of $15$. In order to limit overfitting, we apply the standard regularization techniques of weight decay and dropout~\cite{10.5555/2627435.2670313}. Additionally, batch-normalization~\cite{pmlr-v37-ioffe15} was used in every layer to stabilize training and increase convergence speed. 
Architectural and optimization hyperparameters were estimated using Bayesian optimization with a tree Parzen estimator surrogate function~\cite{NIPS2011_4443} over all inductive datasets. As for the the transductive setting, we employ standard exhaustive search on a predefined hyperparameter grid on \texttt{ogbn-products}, while on \texttt{ogbn-papers100M} we only test a basic configuration that can be found in Supplemetary Materials, along with further details on the hyperparameter search spaces. Given that this dataset represents a directed network, we experimented with operators built via asymmetric normalization of the original directed adjacency matrix and its transpose, as well as their powers.
The \texttt{Wikipedia} dataset, due to the lack of node attributes and labels, is only used to assess scalability: to this end, we randomly generate $100$-dimensional node feature vectors and scalar targets and consider the whole network for both training and inference. No hyperparameter tuning is required in this case. 
%% TODO: add this back as soon as we have a project github link:

\begin{table*}[t]
    \centering
    \caption{Summary of (s)ingle and (m)ulti-label dataset statistics. \texttt{Wikipedia} is used, with random features, for timing purposes only.}%\vspace{-3mm}}
    \label{tab:dataset-statistics-ind}
    \resizebox{\columnwidth}{!}{
    \begin{tabular}{| l | cccccc |}
    \hline
     & $n$ & $|\mathcal{E}|$ & \hspace{-2mm}\textit{Avg. Deg.}\hspace{-2mm} & $d$ & Classes & Train / Val / Test \\
    \hline
    \texttt{Wikipedia}
        & 12,150,976
        & 378,142,420
        & 62
        & 100
        & 2(s)
        & 100\% / --- / 100\% \\
    \texttt{ogbn-papers100M}
        & 111,059,956
        & 1,615,685,872
        & 30
        & 128
        & 172(s)
        & 78\% / 8\% / 14\% \\
    \texttt{ogbn-products}
        & 2,449,029
        & 61,859,140
        & 51
        & 100
        & 47(s)
        & 10\% / 2\% / 88\% \\
    \texttt{Reddit}
        & 232,965
        & 11,606,919
        & 50
        & 602
        & 41(s)
        & 66\% / 10\% / 24\% \\  
    \texttt{Yelp}
        & 716,847
        & 6,977,410
        & 10
        & 300
        & 100(m)
        & 75\% / 10\% / 15\% \\ 
    \texttt{Flickr}
        & 89,250
        & 899,756
        & 10
        & 500
        & 7(s)
        & 50\% / 25\% / 25\% \\  
    \texttt{PPI}
        & 14,755
        & 225,270
        & 15
        & 50
        & 121(m)
        & 66\% / 12\% / 22\% \\
    \hline
    \end{tabular}}
\end{table*}

\begin{table*}[t]
    % \centering
    % \hspace{-1.5cm}
    \caption{Mean and standard deviation of preprocessing, training (one epoch) and inference times, in seconds, on \texttt{ogbn-products} and \texttt{Wikipedia} datasets, computed over 10 runs. {\em SIGN-$r$} denotes architecture with $r$ precomputed operators. Preprocessing and training times for \textit{ClusterGCN} on \texttt{Wikipedia} are not reported due to the clustering algorithm failing to complete.}
    \label{tab:time_stats}
    \resizebox{\columnwidth}{!}{
    \begin{tabular}{| l | ccc | ccc |}
     \hline
     & \multicolumn{3}{c|}{\texttt{ogbn-products}} &
     \multicolumn{3}{c|}{\texttt{Wikipedia}}  \\
    \hline
     & \textit{Preprocessing} & \textit{Training} & \textit{Inference} & \textit{Preprocessing} & \textit{Training} & \textit{Inference} \\ 
    \hline
    \textit{ClusterGCN} &  36.93  $\pm$ 0.52  
                        &   13.34 $\pm$ 0.16
                        &   93.00 $\pm$ 0.68
                        & ---
                        & ---
                        &  183.76 $\pm$ 3.01\\
    \textit{GraphSAINT} &   52.06 $\pm$ 0.54 
                        &    2.89 $\pm$ 0.05
                        &   94.76 $\pm$ 0.81  
                        &  123.60 $\pm$ 1.60
                        &  135.73 $\pm$ 0.06
                        &  209.86 $\pm$ 4.73\\  
    \textit{SIGN-2}     &   88.21 $\pm$ 1.33
                        &    1.04 $\pm$ 0.10
                        &    2.86 $\pm$ 0.10 
                        &  192.88 $\pm$ 0.12
                        &   62.37 $\pm$ 0.17
                        &   13.40 $\pm$ 0.15\\  
    \textit{SIGN-4}     &  160.16 $\pm$ 1.20
                        &    1.54 $\pm$ 0.04
                        &    3.79 $\pm$ 0.08
                        &  326.21 $\pm$ 1.14
                        &   93.84 $\pm$ 0.08
                        &   18.15 $\pm$ 0.05\\  
    \textit{SIGN-6}     &  226.48 $\pm$ 1.43 
                        &    2.05 $\pm$ 0.00
                        &    4.84 $\pm$ 0.08
                        &  459.24 $\pm$ 0.14
                        &  125.24 $\pm$ 0.03
                        &   22.94 $\pm$ 0.02\\  
    \textit{SIGN-8}     &  297.92 $\pm$ 2.92
                        &    2.53 $\pm$ 0.04
                        &    5.88 $\pm$ 0.09
                        &  598.67 $\pm$ 0.82
                        &  154.73 $\pm$ 0.12
                        &   27.69 $\pm$ 0.11\\  
    \hline
    \end{tabular}}
\end{table*}

% {\bf Baselines. \hspace{3mm}}
\paragraph{Baselines}
On the inductive datasets, we compare our method with \textit{GCN}~\cite{kipf2016semi}, \textit{FastGCN}~\cite{fastgcn}, \textit{Stochastic-GCN}~\cite{stochastic-training}, \textit{AS-GCN}~\cite{adaptive-sampling}, \textit{GraphSAGE}~\cite{GraphSAGE}, \textit{ClusterGCN}~\cite{Chiang:2019:CEA:3292500.3330925}, and \textit{GraphSAINT}~\cite{DBLP:journals/corr/abs-1907-04931}, which constitute the current state-of-the-art. On \texttt{ogbn-products} we compare against scalable sampling-free baselines, i.e. a feed-forward network trained over node features only (\textit{MLP}) and on their concatenation with structural \textit{Node2Vec} embeddings~\cite{grover2016node2vec}, and the sampling-based approaches  %\textit{NeighborSampling}~\cite{GraphSAGE},
\textit{ClusterGCN}~\cite{Chiang:2019:CEA:3292500.3330925} and \textit{GraphSAINT}~\cite{DBLP:journals/corr/abs-1907-04931}. 
As for \texttt{ogbn-papers100M}, \textit{SIGN} is compared with sampling-free baselines: an \textit{MLP} trained on node features and an \textit{S-GCN} model. Sampling-based methods have not been scaled yet to this benchmark. All results for OGB datasets are directly taken from the latest version of the arXiv paper~\cite{ogb2020} (\emph{v4}, at the time of writing). Lastly, being \textit{S-GCN} an important baseline for our model, we additionally report its performance on all the other datasets as well. In this case we choose power $L$ of its (only) operator $\mathbf{A}^L$ as the value $p$ of the best corresponding \textit{SIGN($p$,$s$,$t$)} configuration and we tune its hyperparameters in the same space searched for \textit{SIGN}.

% {\bf Implementation. \hspace{3mm}}
\paragraph{Implementation}
\textit{SIGN} is implemented using Pytorch~\cite{NEURIPS2019_9015}.
All experiments, including timings, were run on an AWS p2.8xlarge instance, with 8 NVIDIA K80 GPUs, 32 vCPUs, a processor Intel(R) Xeon(R) CPU E5-2686 v4 @ 2.30GHz and 488GiB of RAM. 

\subsection{Results}

\begin{table*}[t]
    %\hspace{-1cm}
    \caption{Micro-averaged F1 scores. 
    %average and standard deviation over 10 runs with the same train/val/test split but different random model initialization.
    For \textit{SIGN}, we show the best performing configurations. 
    The top three performance scores are highlighted as: {\bf \bf \color{red} First}, {\bf \bf \color{violet} Second}, {\bf Third}. 
    }
    \label{tab:results_inductive}
    \centering
        \begin{tabular}{| l | cccc |}
        \hline
         & \texttt{Reddit} & \texttt{Flickr} & \texttt{PPI} & \texttt{Yelp} \\ \hline
        \textit{GCN}~\cite{kipf2016semi} & 0.933$\pm$0.000 & 0.492$\pm$0.003 & 0.515$\pm$0.006 & 0.378$\pm$0.001\\
        \textit{FastGCN}~\cite{fastgcn} & 0.924$\pm$0.001 &  {\bf 0.504$\pm$0.001} & 0.513$\pm$0.032 & 0.265$\pm$0.053\\
        \textit{Stochastic-GCN}~\cite{stochastic-training} & {\bf 0.964$\pm$0.001} &  0.482$\pm$0.003 & {\bf 0.963$\pm$0.010} & {\bf \color{violet} 0.640$\pm$0.002}\\
        \textit{AS-GCN}~\cite{adaptive-sampling} & 0.958$\pm$0.001 &  {\bf 0.504$\pm$0.002} & 0.687$\pm$0.012 & ---\\
        \textit{GraphSAGE}~\cite{GraphSAGE} & 0.953$\pm$0.001 &  0.501$\pm$0.013 & 0.637$\pm$0.006 & \textbf{0.634$\pm$0.006}\\ 
        \textit{ClusterGCN}~\cite{Chiang:2019:CEA:3292500.3330925} & 0.954$\pm$0.001 &  0.481$\pm$0.005 & 0.875$\pm$0.004 & 0.609$\pm$0.005 \\ 
        \textit{GraphSAINT}~\cite{DBLP:journals/corr/abs-1907-04931} & {\bf \color{violet} 0.966$\pm$0.001} & {\bf \color{violet} 0.511$\pm$0.001} & {\bf \color{red} 0.981$\pm$0.004} & {\bf \color{red} 0.653$\pm$0.003} \\
    %    \hline
    %    \textit{\textbf{SIGN (Ours)}} & {\bf \color{violet} 0.966$\pm$0.003} & 0.503$\pm$0.003 & {\bf \color{black} 0.965$\pm$0.002} & 0.623$\pm$0.005\\
%        \textit{\textbf{SIGN (Extra Ops)}} & {\bf \color{red} 0.968$\pm$0.003} & {\bf \color{violet} 0.508$\pm$0.001} & {\bf \color{violet} 0.970$\pm$0.003} & 0.625$\pm$0.003\\
%    %    0.774$\pm$ 0.002}\\
%    
%        \textit{\textbf{SIGN}} & {\bf \color{violet} 0.966$\pm$0.003} & 0.503$\pm$0.003 & {\bf \color{black} 0.965$\pm$0.002} & 0.623$\pm$0.005\\
        %\textit{\textbf{SIGN}} & {\bf \color{red} 0.969$\pm$0.003} & {\bf \color{red} 0.514$\pm$0.001} & {\bf \color{violet} 0.970$\pm$0.003} & 0.625$\pm$0.003\\
        \textit{S-GCN}~\cite{pmlr-v97-wu19e} & 0.949$\pm$0.000 & 0.502$\pm$0.001 & 0.892$\pm$0.015 & 0.358$\pm$0.006\\% here yelp is run [24] of the tuning
        \hline
        \textit{SIGN} & {\bf \color{red} 0.968$\pm$0.000} & {\bf \color{red} 0.514$\pm$0.001} & {\bf \color{violet} 0.970$\pm$0.003} & 0.631$\pm$0.003\\%0.625$\pm$0.003\\
    %    0.774$\pm$ 0.002}\\
        %$(p,s,t)$ & $(3,3,0)$ & $(4,0,1)$ & $(2,0,1)$ & $(2,0,1)$ \\
        $(p,s,t)$ & $(4,2,0)$ & $(4,0,1)$ & $(2,0,1)$ & $(2,0,1)$ \\
    %    0.774$\pm$ 0.002}\\

        \hline
    \end{tabular} 
\end{table*}

% {\bf Inductive. \hspace{3mm}}
\paragraph{Inductive}
Table \ref{tab:results_inductive} presents the results on the inductive dataset. In line with \cite{DBLP:journals/corr/abs-1907-04931}, we report the micro-averaged F1 score means and standard deviations computed over 10 runs. For each dataset we report the best performing \textit{SIGN} configuration, specifying the maximum power for each of the three employed operators.
\textit{SIGN} outperforms other methods on \texttt{Reddit} and \texttt{Flickr}, and performs competitively to state-of-the-art on \texttt{PPI}. Our performance on \texttt{Yelp} is worse than in the other datasets; we hypothesize that a more tailored choice of operators is required to better suit the characteristics of this dataset.
Interestingly, \textit{SIGN} significantly outperforms \textit{S-GCN} in all datasets, suggesting that the additional expressivity introduced by the different operators in our model is required for effective learning.

% {\bf Transductive.\hspace{3mm}} 
\paragraph{Transductive}
\textit{SIGN} obtains state-of-the-art results on the \newline \texttt{ogbn-papers100M} dataset (Table \ref{tab:ogbn-100m}), outperforming other sampling-free methods by at least $1.8\%$. This shows that \textit{SIGN} can scale to massive graphs while retaining ample expressivity. Sampling based methods have not been scaled yet to this benchmark. In \texttt{ogbn-papers100M} only $ \sim 1.35\%$ of nodes are labeled; at \emph{each} training and inference iteration these methods still need to perform computation on subgraphs where the majority of nodes are unlabeled and thus do not contribute to the computation of loss and evaluation metrics. On the contrary, at training and inference \textit{SIGN} only processes the required labeled nodes given that the graph has already been employed during the one-time pre-computation phase, thus avoiding this redundant computation and memory usage.

\texttt{ogbn-products} results are reported in Table \ref{tab:results_ogb}.
%This benchmark is particularly hard as the training/validation/test split is not random, but rather reflects a more realistic and challenging scenario where labels are first assigned to important nodes (according to sales ranking) and models are subsequently used to make predictions on less important ones. In order to test the capability of models to generalize to out-of-distribution node, in accordance to \cite{ogb2020} we report performance for all models on training, validation, and test sets.
\textit{SIGN} outperforms all other sampling-free methods by at least $2.7\%$. However, contrary to the inductive benchmarks, sampling methods outperform \textit{SIGN} and appear to generally be more suitable to this dataset. We hypothesise that, on this particular task, sampling may implicitly act as a regularizer, making these methods generalize better to the held-out test set, which in this dataset is sampled from a different distribution w.r.t. training and validation nodes \cite{ogb2020}. This phenomenon, as well as its connection to the DropEdge method~\cite{rong2019dropedge} and the bottleneck problem~\cite{alon2020bottleneck}, would be object of further investigation. 

% {\bf Runtime. \hspace{3mm}}
\paragraph{Runtime}
While performing on par or better than state-of-the-art methods on most benchmarks in terms of accuracy, our method has the advantage of being significantly faster than other methods for large graphs. 
We perform comprehensive timing comparisons on \texttt{ogbn-products} and \texttt{Wikipedia} datasets and report average training, inference, and preprocessing times in Table~\ref{tab:time_stats}. For these experiments, we run the implementations of \textit{ClusterGCN} and \textit{GraphSAINT} provided in the OGB code repository\footnote{https://github.com/snap-stanford/ogb/tree/master/examples/nodeproppred/products}. 

We use these datasets rather than \texttt{ogbn-papers100M} so we can compare to \textit{ClusterGCN} and \textit{GraphSAINT}. For completeness we report, however, that on \texttt{ogbn-papers100M} the best performing \textit{SIGN(3,3,3)} model completes one evaluation pass on the validation set in $1.99\pm0.05$ seconds and on the test set in $3.34\pm0.04$ seconds (statistics are estimated over $10$ runs and include the time required by device data transfers and by the computation of evalution metric).

\begin{figure}
    \centering
    \includegraphics[width=0.5\linewidth]{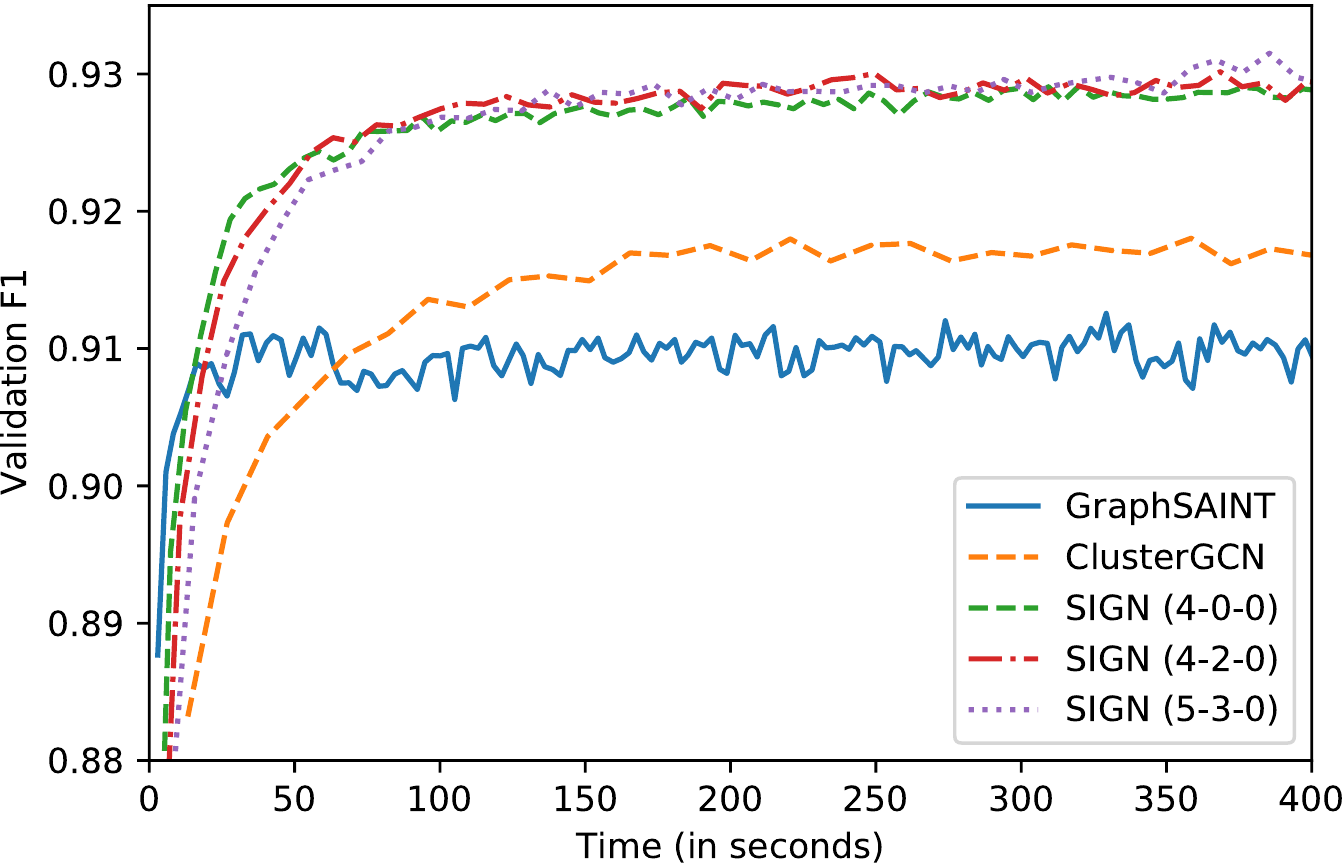}
    \caption{Convergence of different methods on \texttt{ogbn-products}.}
    \label{fig:ogb_timings}
\end{figure}

Our model is faster than \textit{ClusterGCN} and of comparable speed w.r.t. \textit{GraphSAINT} in training\footnote{Traning time is measured as forward-backward time to complete one epoch.}, while being by far the fastest approach in inference: all \textit{SIGN} architectures are always at least one order of magnitude faster than other methods, with the largest one ($8$ operators) requiring no more than $30$ seconds to perform inference on over $12$M nodes.
\textit{SIGN}'s preprocessing is slightly longer than other methods, but we notice that most of the calculations %, including the approximation of PPR diffusion matrices, 
can be cast as sparse matrix multiplications and easily parallelized with frameworks for distributed computing. We envision to engineer faster and even more scalable \textit{SIGN} preprocessing implementations in future developments of this work.
Finally, in order to also study the convergence behavior of our proposed model, in Figure~\ref{fig:ogb_timings} we plot the validation performance on \texttt{ogbn-products} from the start of the training as a function of run time for \textit{ClusterGCN}, \textit{GraphSAINT} and several \textit{SIGN} configurations. We observe that \textit{SIGN} does not only converge to a better validation accuracy than other methods, but also exhibits much faster convergence than \textit{ClusterGCN} and comparable speed than to \textit{GraphSAINT}.

\begin{table}[t]
    \centering
    \caption{Performance on \texttt{ogbn-products}. {\em SIGN($p$,$s$,$t$)} refers to a configuration using $p$, $s$, and $t$ powers of simple, PPR-based, and triangle-based adjacency matrices. The top three performance scores are highlighted as: {\bf \color{red} First}, {\bf \color{violet} Second}, {\bf Third}.}
    \label{tab:results_ogb}
    \begin{tabular}{| l | ccc |}
        \hline
        & \textit{Training} & \textit{Validation} & \textit{Test} \\
        \hline
        \textit{MLP} & 84.03$\pm$0.93 & 75.54$\pm$0.14 & 61.06$\pm$0.08 \\
        \textit{Node2Vec} & 93.39$\pm$0.10 & 90.32$\pm$0.06 & 72.49$\pm$0.10 \\
        \textit{S-GCN}($L$=5) & 92.54$\pm$0.09 & 91.38$\pm$0.07 & 74.87$\pm$0.25 \\
        \hline
  %      \textit{NeighborSampling} & 92.96$\pm$0.0 & 91.70$\pm$0.09 & 78.70$\pm$0.3 \\
        \textit{ClusterGCN} & 93.75$\pm$0.13 & 92.12$\pm$0.09 & {\bf \color{violet} 78.97$\pm$0.33} \\ 
        \textit{GraphSAINT} & 92.71$\pm$0.14 & 91.62$\pm$0.08 & {\bf \color{red} 79.08$\pm$0.24} \\ 
        \hline
        \textit{SIGN}(3,0,0) & 96.21$\pm$0.31 & {\bf 92.99$\pm$0.05} & 76.52$\pm$0.14 \\
        \textit{SIGN}(3,0,1) & {\bf 96.46$\pm$0.29} & 92.93$\pm$0.04 & 75.73$\pm$0.20 \\
        \textit{SIGN}(3,3,0) & {\bf \color{violet} 96.87$\pm$0.23} & {\bf \color{violet} 93.02$\pm$0.04} & 77.13$\pm$0.10 \\
        \textit{SIGN}(5,0,0) & 95.99$\pm$0.69 & 92.98$\pm$0.18 & 76.83$\pm$0.39 \\
        \textit{SIGN}(5,3,0) & {\bf \color{red} 96.92$\pm$0.46} & {\bf \color{red} 93.10$\pm$0.08} & {\bf  77.60$\pm$0.13}\\
        \hline
    \end{tabular}
\end{table}

% {\bf Ablation study. \hspace{3mm}}
\paragraph{Ablation study}
How do different operator combinations affect \textit{SIGN} performance? Results obtained with different choices of operators and their powers are reported in Tables \ref{tab:results_ogb}, \ref{tab:ogbn-100m} and~\ref{tab:inductive_ablation} for, respectively, the the transductive \texttt{ogbn-products} and \texttt{ogbn-papers100M} and inductive datasets. We notice that best performance is obtained on each benchmark by a specific combination of operators, remarking the fact that each dataset features particular topological and content characteristics requiring suitable filters. Interestingly, we also observe that while PPR operators do not bring significant improvements in the inductive setting (being even harmful in certain cases), they are beneficial on the transductive \texttt{ogbn-products}. This finding is in accordance with \cite{klicpera_diffusion_2019}, where the effectiveness of PPR diffusion operators in transductive settings has been extensively studied. Finally, we notice promising results attained in \texttt{Flickr} and \texttt{PPI} inductive settings by pairing standard adjacency matrices with a triangle-induced one.
Studying the effect of operators induced by more complex network motifs is left for future research.

\begin{table*}[t]
    \centering
    \caption{Impact of various operator combinations on inductive datasets. Best results are in bold.}
    \label{tab:inductive_ablation}
    \begin{tabular}{| c | cccc |}
        \hline
        & \texttt{Reddit} & \texttt{Flickr} & \texttt{PPI} & \texttt{Yelp} \\
        \hline
        \textit{SIGN}(2,0,0) & 0.966$\pm$0.003 & 0.503$\pm$0.003 & 0.965$\pm$0.002 & 0.623$\pm$0.005 \\
        \textit{SIGN}(2,0,1) & 0.966$\pm$0.000 & 0.510$\pm$0.001 & \textbf{0.970$\pm$0.003} & \textbf{0.631$\pm$0.003} \\%\textbf{0.625$\pm$0.003} \\%0.622$\pm$0.004 \\
        \textit{SIGN}(2,2,0) & 0.967$\pm$0.000 & 0.495$\pm$0.002 & 0.964$\pm$0.003 & 0.617$\pm$0.005 \\
        \textit{SIGN}(4,0,0) & 0.967$\pm$0.000 & 0.508$\pm$0.001 & 0.959$\pm$0.002 & 0.623$\pm$0.004 \\
        \textit{SIGN}(4,0,1) & 0.967$\pm$0.000 & \textbf{0.514$\pm$0.001} & 0.965$\pm$0.003 & 0.623$\pm$0.004 \\%0.622$\pm$0.003 \\
        \textit{SIGN}(4,2,0) & \textbf{0.968$\pm$0.000} & 0.500$\pm$0.001 & 0.930$\pm$0.010 & 0.618$\pm$0.004 \\
        \textit{SIGN}(4,2,1) & 0.967$\pm$0.000 & 0.508$\pm$0.002 & 0.969$\pm$0.001 & 0.620$\pm$0.004 \\%0.616$\pm$0.005 \\
        \hline
    \end{tabular} 
\end{table*}

\begin{table}[t]
    \centering
    \caption{Results on \texttt{ogbn-papers100M}, the largest public graph dataset with over 110 million nodes. {\em SIGN($p$,$d$,$f$)} refers to a configuration using $p$, $d$, and $f$ powers of simple undirected, directed and directed-transposed adjacency matrices. The top three performance scores are highlighted as: {\bf \color{red} First}, {\bf \color{violet} Second}, {\bf Third}.}
    \label{tab:ogbn-100m}
    \begin{tabular}{| l | ccc |}
        \hline
        & \textit{Training} & \textit{Validation} & \textit{Test} \\
        \hline
        \textit{MLP}
            & 54.84$\pm$0.43
            & 49.60$\pm$0.29
            & 47.24$\pm$0.31 \\
        \textit{Node2Vec}
            & ---
            & 55.60$\pm$0.23
            & 58.07$\pm$0.28 \\
        \textit{S-GCN}($L$=3)
            & 67.54$\pm$0.43
            & 66.48$\pm$0.20
            & 63.29$\pm$0.19 \\
        \hline
        \textit{SIGN}(3,0,0)
            & \textbf{70.18$\pm$0.37}
            & \textbf{67.57$\pm$0.14}
            & \textbf{64.28$\pm$0.14}\\
        \textit{SIGN}(3,1,1)
            & \textbf{\color{violet} 72.24$\pm$0.32}
            & \textbf{\color{violet} 67.76$\pm$0.09}
            & \textbf{\color{violet} 64.39$\pm$0.18 }\\
        \textit{SIGN}(3,3,3)
            & \textbf{\color{red} 73.94$\pm$0.72}
            & \textbf{\color{red} 68.6$\pm$0.04}
            & \textbf{\color{red} 65.11$\pm$0.14} \\
        \hline
    \end{tabular} 
\end{table}

\section{Conclusion and Future Work} \label{sec:conclusion}
 
In this paper we presented \textit{SIGN}, a sampling-free Graph Neural Network model that is able to easily scale to gigantic graphs while retaining enough expressive power to attain competitive results in all large-scale graph learning benchmarks. \textit{SIGN} attains state-of-the-art results on many of them, including the massive \texttt{ogbn-papers100M}, currently the largest  publicly available node-classification benchmark  with $ \sim111$M nodes and $ \sim1.6$B edges.
Our experiments have further demonstrated that the ability of our model to flexibly incorporate diverse, possibly domain-specific, operators is crucial to overcome the expressivity limitations of other sampling-free scalable models such as \textit{S-GCN}, which \textit{SIGN} has constantly outperformed over all datasets.
Overall, our architecture achieves an optimal trade-off between simplicity and expressiveness; as it has shown to attain competitive results with fast training and inference, it represents the most suitable architecture for scalable applications to web-scale graphs.

% {\bf Depth vs width for graph neural networks. \hspace{3mm}}
\paragraph{Depth vs. width for Graph Neural Networks}
Our results have shown that it is possible to obtain competitive -- and often state-of-the-art -- results with one single graph convolutional layer and hence a shallow architecture. An important question is, therefore, when one should apply deep architectures to graphs, where by `depth' we refer to the number of stacked graph convolutional layers.
Deep Graph Neural Networks are notoriously hard to train due to vanishing gradients and feature smoothing \cite{li2018adaptive,klicpera2018predict,wu2020comprehensive}, and, although recent works have shown that these issues can be addressed to some extent \cite{jk,gong2020geometrically, li2019deepgcns, zhao2019pairnorm, rong2019dropedge}, yet extensive experiments conducted in \cite{rong2019dropedge} showed that depth often does not bring any significant gain in performance w.r.t. to shallow baselines.
A promising direction for future investigation is, rather than `going deep', to `go wide', in the sense of exploring more expressive local operators. We believe this to be especially crucial in all those settings where scalability is a concern of paramount importance, such as in industrial large-scale systems.

% {\bf Extensions. \hspace{3mm}}
\paragraph{Extensions}
In our experiments, triangle-based operators showed promising results. Possible extensions can employ operators that account for higher-order structures such as simplicial complexes \cite{barbarossa2019topological}, paths \cite{flam2020neural}, or motifs \cite{monti2018motifnet} that can be tailored to the specific problem. 
Furthermore, temporal information can be integrated e.g. in the form of temporal motifs \cite{paranjape2017motifs}. 

% {\bf Limitations. \hspace{3mm}}
\paragraph{Limitations}
While our method relies on linear graph aggregation operations of the form $\mathbf{B}\mathbf{X}$ for efficient precomputation, it is possible to make the diffusion operator dependent on the node features (and edge features, if available). 
In particular, graph attention \cite{DBLP:conf/iclr/VelickovicCCRLB18} and similar mechanisms \cite{Monti2016GeometricDL} use $\mathbf{B}_{\boldsymbol{\theta}}(\mathbf{X})$, where $\boldsymbol{\theta}$ are learnable parameters. The limitation is that such operators preclude efficient precomputation, which is key to the efficiency of our approach. 
Attention can be implemented in our scheme by training on a small subset of the graph to first determine the attention parameters, then fixing them to precompute the diffusion operator that is used during training and inference.  
%More generally, architectures with multiple convolutional layers are achievable by layer-wise training. 

%\bibliographystyle{unsrt}
\bibliographystyle{ACM-Reference-Format}
\bibliography{neurips_2020}

%%
%% If your work has an appendix, this is the place to put it.

\newpage
\appendix

\section{Datasets}

\subsection{Inductive datasets}

\texttt{Reddit} \cite{GraphSAGE} and \texttt{Flickr} \cite{DBLP:journals/corr/abs-1907-04931} are multiclass classification problems, \texttt{Yelp} \cite{DBLP:journals/corr/abs-1907-04931} and \texttt{PPI} \cite{Zitnik_2017} are multilabel classification instances. In \texttt{Reddit}, the task is to predict communities of online posts based on user comments. In \texttt{Flickr} the task is image categorization based on the description and common properties of online images. In \texttt{Yelp} the objective is to predict business attributes based on customer reviews; the task of \texttt{PPI} consists in predicting protein functions from the interactions of human tissue proteins. Further details on the generation of the \texttt{Yelp} and \texttt{Flickr} datasets can be found in \cite{DBLP:journals/corr/abs-1907-04931}.

\subsection{Transductive dataset}

\texttt{ogbn-products} \cite{ogb2020} represents an Amazon product co-purchasing network \cite{Bhatia16} where the task is to predict the category of a product in a multi-class classification setup. Dataset splitting is not random, sales ranking (popularity) is instead used to split nodes into training/validation/test. Top 10\% products in the ranking are assigned to the training set, next top 2\% to validation and the remaining 88\% of products are for testing.

\texttt{ogbn-papers100M} \cite{ogb2020} represents a directed citation network of $ \sim $ 111 million academic papers, where the task is to leverage information from the entire citation network to infer the labels (subject areas) of a smaller subset of ArXiv papers. The splitting strategy is time-based. Specifically, the training nodes (with labels) are all ArXiv papers published until 2017, validation nodes are ArXiv papers published in 2018 and test nodes are ArXiv papers published since 2019.

\subsection{Wikipedia}

\texttt{Wikipedia} links is a large-scale directed network of links between articles in the English version of Wikipedia. For the sake of our timing experiments the network has been turned into undirected. Node features have been randomly generated with a dimensionality of $100$ as in \texttt{ogbn-products}.

\section{Model Selection and Hyperparameter Tuning}

Tuning involved the following architectural and optimization hyperparameters: weight decay, dropout rate, batch size, learning rate, number of feedforward layers and units both in inception and classification modules. For each inductive experiment we chose the set of hyperparameters matching the best average validation loss calculated over $5$ runs. For the the transductive setting we kept, instead, the set of hyperparameters with minimum validation loss over a single run. The hyperparameter search space for the inductive setting and grid for the transductive one are described in Table \ref{tab:hyperparameter_search}. The estimated hyperparameters for each best \textit{SIGN} configuration are reported in Table \ref{tab:hyperparameter_tuned} for inductive datasets and Table \ref{tab:hyperparameter_tuned_ogb} for the transductive ones.

\begin{table*}[h!]
    \caption{Hyperparameter search space/grid. Ranges in the form [\textit{low}, \textit{high}] and sampling distributions. \textit{Inception Layers} and \textit{Classification Layers} are the number of feedforward layers in the representation part of the model (replacing $\Theta$) and the classification part of the model (replacing $\Omega$) respectively. The only exception is represented by \textit{Yelp}, for which the $\Omega$ module was kept shallow (no hidden layers) to allow for lighter training and the left bounds on the dropout, learning rate and batch size intervals were lowered to, respectively, $0.0$, $0.00001$ and $60$.}%as even mild values for this hyperparameter showed to be particularly detrimental in early experiments.}
    \label{tab:hyperparameter_search}
    \vskip 0.15in
    \begin{center}
    \begin{small}
    \begin{sc}
    \begin{tabular}{| r | c | cc |}
    \hline
    & \multicolumn{1}{c|}{Transductive} & \multicolumn{2}{c|}{Inductive} \\
    \hline
    Hyperparameter & Values & Space & Distribution \\
    \hline
    \textit{Learning Rate}
        & 0.0001, 0.001 
        & [0.0001, 0.0025]
        & Uniform \\
    \textit{Batch Size}
        & 4096, 8192, 16384
        & [128, 2048]
        & Quantized Uniform \\
    \textit{Dropout}
        & 0.5
        & [0.2, 0.8]
        & Uniform \\ 
    \textit{Weight Decay}
        & 0.0, 0.00001
        & [0, 0.0001]
        & Uniform \\
    \textit{Inception Layers}
        & 1
        & 1, 2
        & ---\\ 
    \textit{Inception Units}
        & 256, 512
        & [128, 512]
        & Quantized Uniform\\ 
    \textit{Classification Layers}
        & 1
        & 1, 2
        & --- \\
    \textit{Classification Units}
        & 256, 512
        & [512, 1024]
        & Quantized Uniform\\ 
    \textit{Activation}
        & PReLU
        & ReLU, PReLU
        & --- \\
    \hline
    \end{tabular}
    \end{sc}
    \end{small}
    \end{center}
    \vskip -0.1in
\end{table*}

\begin{table*}[h!]
    \caption{Hyperparameters chosen for the best configuration of SIGN on inductive datasets.}
    \label{tab:hyperparameter_tuned}
    \vskip 0.15in
    \begin{center}
    \begin{small}
    \begin{sc}
    \resizebox{\columnwidth}{!}{
    \begin{tabular}{| r | cccc |}
    \hline
    Hyperparameter & Reddit & Flickr & PPI & Yelp\\% & OGBN-Products \\
    \hline
    \textit{Learning Rate}
        & 0.00012278578238312588
        & 0.0017230142114465549
        & 0.0014386686616183625
        & 0.00005\\
    \textit{Dropout}
        & 0.707328910934901
        & 0.7608352140584778
        & 0.3085607444207686
        & 0.05\\
    \textit{Weight Decay}
        & 9.176773905054599e-05
        & 9.419820474221673e-05
        & 3.2571631135664696e-06
        & 4.452466189193362e-07\\
    \textit{Batch Size}
        & 830
        & 330
        & 210
        & 90\\
    \textit{Inception Layers}
        & 1
        & 2
        & 2
        & 2\\
    \textit{Inception Units}
        & 460
        & 465
        & 315
        & 320\\
    \textit{Classification Layers}
        & 1
        & 1
        & 2
        & 0\\
    \textit{Classification Units}
        & 675
        & 925
        & 870
        & ---\\
    \textit{Activation}
        & ReLU
        & PReLU
        & ReLU
        & ReLU\\
    \hline
    \end{tabular}}
    \end{sc}
    \end{small}
    \end{center}
    \vskip -0.1in
\end{table*}

\begin{table*}[h!]
    \caption{Hyperparameters chosen for the best configuration of SIGN on \texttt{ogbn-product} and those used on the \texttt{ogbn-product} dataset.}
    \label{tab:hyperparameter_tuned_ogb}
    \vskip 0.15in
    \begin{center}
    \begin{small}
    \begin{sc}
    \begin{tabular}{| r | c | c |}
    \hline
    Hyperparameter & \texttt{ogbn-products} & \texttt{ogbn-papers100M} \\
    \hline
    \textit{Learning Rate}
        & 0.0001
        & 0.001 \\
    \textit{Dropout}
        & 0.5
        & 0.1 \\
    \textit{Weight Decay}
        & 0.0001
        & 0.0 \\
    \textit{Batch Size}
        & 4096
        & 256 \\
    \textit{Inception Layers}
        & 1
        & 1 \\
    \textit{Inception Units}
        & 512
        & 256 \\
    \textit{Classification Layers}
        & 1
        & 3 \\
    \textit{Classification Units}
        & 512 
        & 256 \\
    \textit{Activation}
        & PReLU
        & ReLU \\
    \hline
    \end{tabular}
    \end{sc}
    \end{small}
    \end{center}
    \vskip -0.1in
\end{table*}

\section{Triangle-based Operators}

The triangle operator encodes the concept of \emph{homophily} with a stronger acceptation with respect to the adjacency matrix: two nodes are connected by an edge only if they are both part of the same closed triad, i.e. if they are connected together and are both connected to the same node. Edge weights are proportional to the amount of triangles an edge belongs to, and they are normalised row-wise so to represent, for each node in a neighbourhood, its relative importance with respect to all the other neighbors.

This brings us to two considerations: first of all, the triangle operator is not carrying information related to nodes which were not already in the neighborhood. Secondly, it emphasizes the connections with those neighbors which are more related to our source node in virtue of the relationship described above. We can thus envision this operator being more useful in those graphs where this kind of relationship can be more discriminative within a neigborhood.

\begin{figure}
    \begin{subfigure}[t]{0.465\linewidth}
    \centering
    \includegraphics[width=\textwidth]{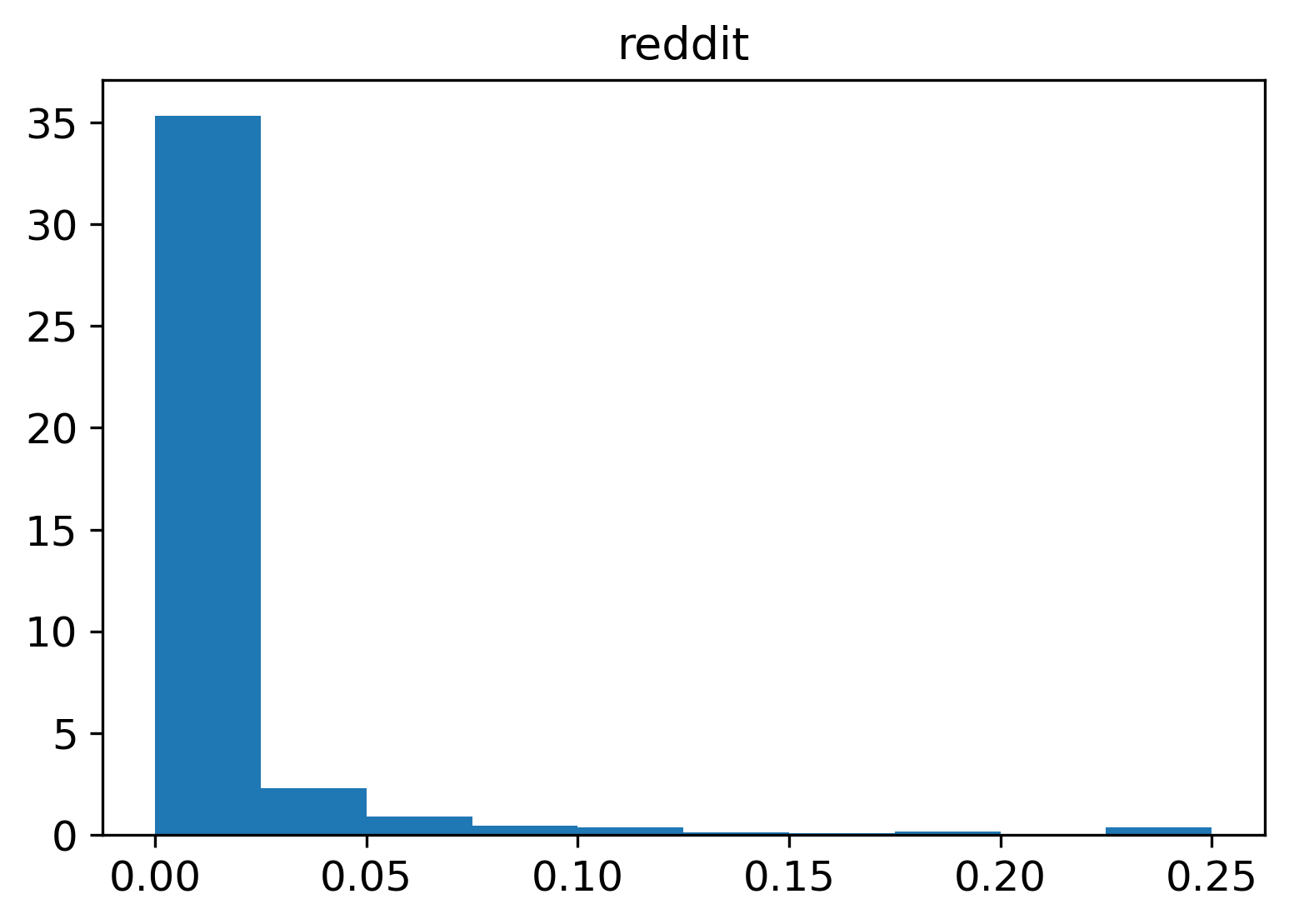}
    \end{subfigure}\hfill%
    \begin{subfigure}[t]{0.465\linewidth}
    \centering
    \includegraphics[width=\linewidth]{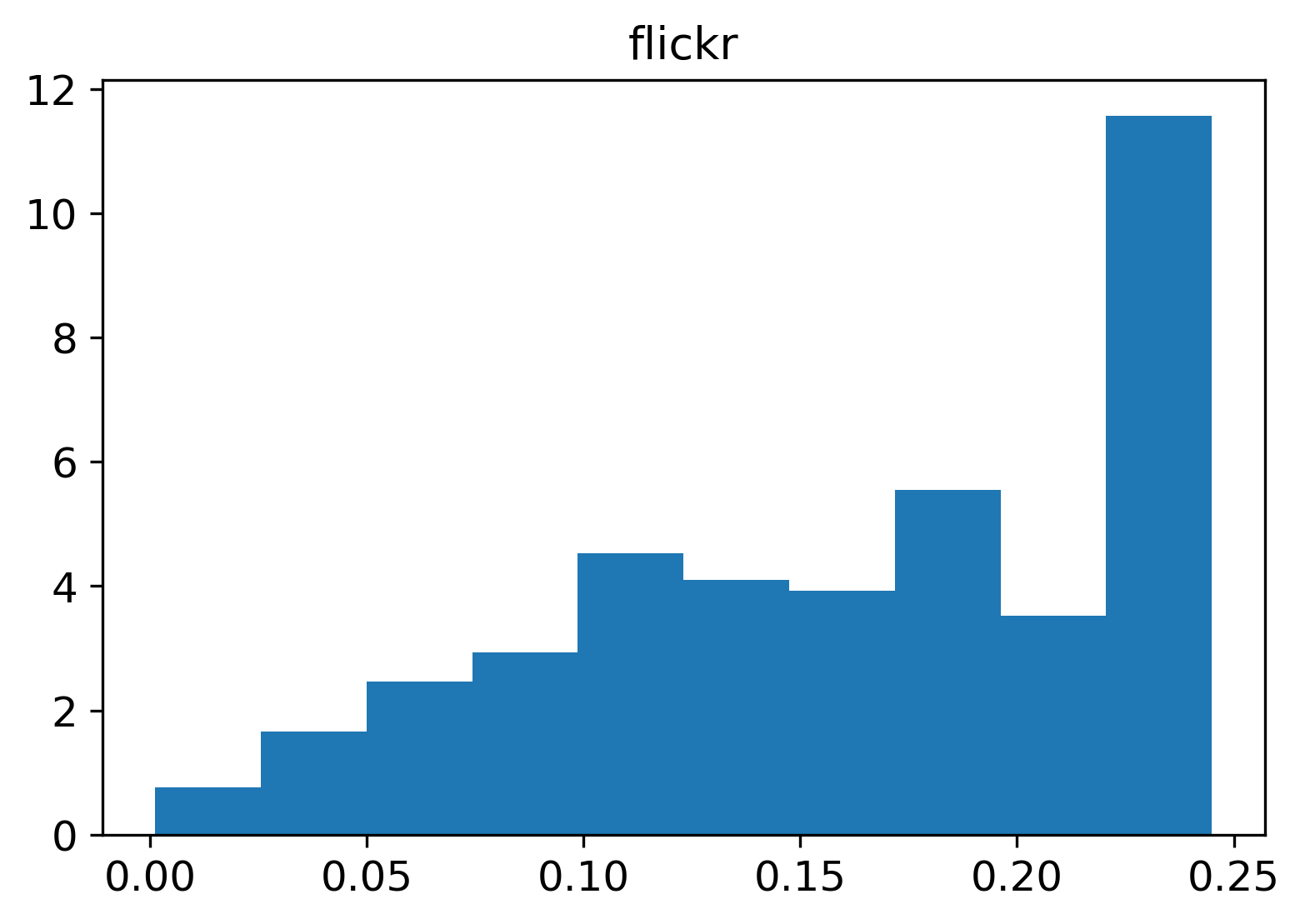}
    \end{subfigure}\hfill%
    \begin{subfigure}[t]{0.465\linewidth}
    \centering
    \includegraphics[width=\linewidth]{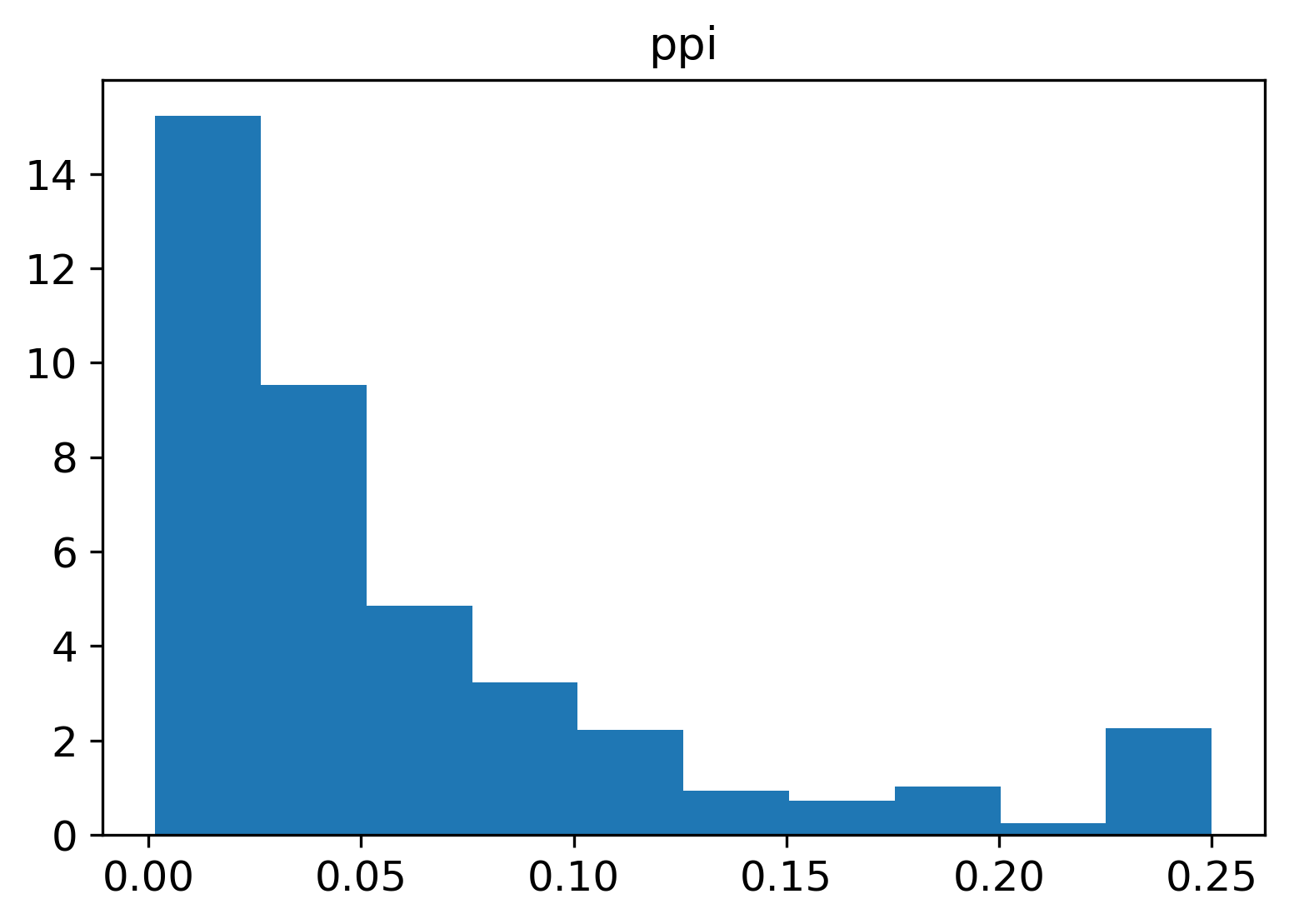}
    \end{subfigure}\hfill%
    \begin{subfigure}[t]{0.465\linewidth}
    \centering
    \includegraphics[width=\textwidth]{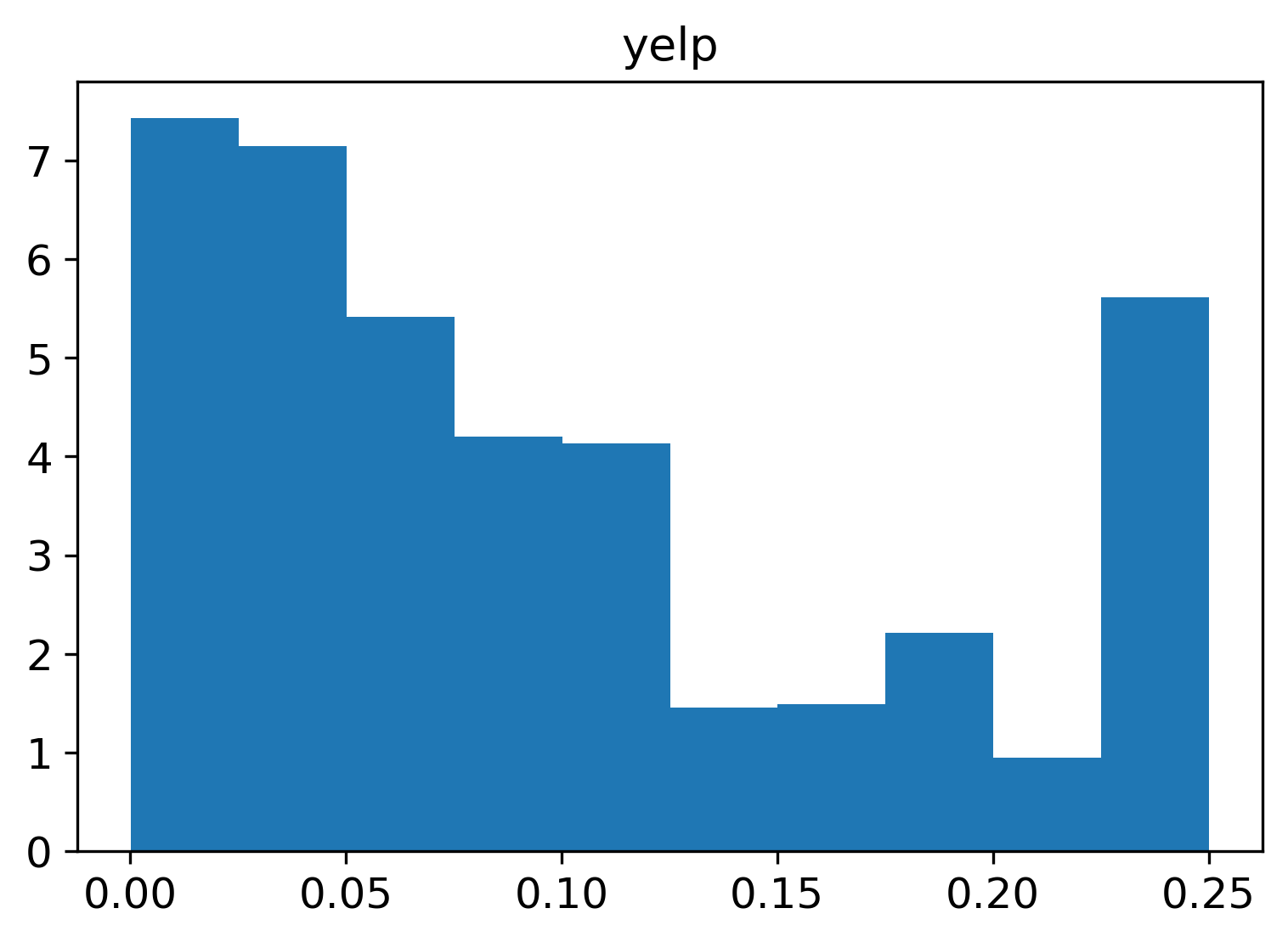}
    \end{subfigure}
    \caption{Normalized frequency distributions for row-wise variations on the diffusion weights of triangle operators over inductive datasets. Variations are measured as the standard deviation on the weight value over original neighborhoods from the test graph.}
    \label{fig:triangles}
\end{figure}

To verify this, in Figure~\ref{fig:triangles} we plot the normalized frequency distribution of intra-neighborhood standard deviation for the weights of triangle operators. It is interesting to notice the significantly different trends characterizing \texttt{Flickr} and \texttt{Reddit}, two datasets where triangle operators have experimentally brought, respectively, relative large and small performance improvement. \texttt{Flickr} tends to exhibit larger weight variations than other datasets, while, on the contrary, \texttt{Reddit} is the dataset where the smallest intra-neighborhood variation is observed. This suggests how, in \texttt{Flickr}, the triangle operator is able to restrict feature aggregation to a subset of the original neighbors --those co-occurring in the larger number of triangles-- while in \texttt{Reddit} it mostly boils down to uniform averaging, making this operator not much more expressive than a simple adjacency matrix.

For replicability we report that, in the computation of triangle operators for \texttt{PPI}, %and \textit{Yelp}, 
we retained the self-loops already present in the original dataset. %s.
Investigations on how the presence of these edges affects the expressiveness of the triangle operator are left for future work.

\end{document}